\def\year{2021}\relax
\def\@fnsymbol#1{\ensuremath{\ifcase#1\or *\or \dagger\or \ddagger\or
   \mathsection\or \mathparagraph\or \|\or **\or \dagger\dagger
   \or \ddagger\ddagger \else\@ctrerr\fi}}
\documentclass[letterpaper]{article} 
\usepackage{aaai21}  
\usepackage{times}  
\usepackage{helvet} 
\usepackage{courier}  
\usepackage[hyphens]{url}  
\usepackage{graphicx} 
\usepackage{natbib}  
\usepackage{caption} 
\usepackage{booktabs}
\usepackage{graphicx}
\usepackage{multirow} 
\usepackage{tabularx}
\usepackage{textcomp}
\usepackage{fourier} 
\usepackage{array}
\usepackage{makecell}
\usepackage{xcolor}
\usepackage{enumerate}
\usepackage{amsmath}
\usepackage{hyperref}

\usepackage[T1]{fontenc}
\usepackage{babel}

\usepackage{subcaption,siunitx}
\usepackage[margin=1in]{geometry} 
\newcommand{\oneS}{\ensuremath{{}^{\textstyle *}} }

\newcommand{\ssymbol}[1]{^{\@fnsymbol{#1}}}
\newcommand{\rgdiff}{$\text{RG}_{\text{diff}}$}
\newcommand{\fdiff}{$\text{F}_{\text{diff}}$}

\title{On Generating Extended Summaries of Long Documents}
\author{
    Sajad Sotudeh$\ssymbol{2}$, Arman Cohan$\ssymbol{3}$, Nazli Goharian$\ssymbol{2}$\\

}
\affiliations{
    $\ssymbol{2}$ IR Lab, Georgetown University, Washington D.C., USA \\
    \texttt{\{sajad, nazli\}@ir.cs.georgetown.edu}, \\
    
    $\ssymbol{3}$ Allen Institute for Artificial, Seattle, WA, USA \\
    \texttt{armanc@allenai.org}

}

\begin{document}

\maketitle
 
\begin{abstract}
Prior work in document summarization has mainly focused on generating short summaries of a document. While this type of summary helps get a high-level view of a given document, it is desirable in some cases to know more detailed information about its salient points that can't fit in a short summary. This is typically the case for longer documents such as a research paper, legal document, or a book. In this paper, we present a new method for generating extended summaries of long papers. Our method exploits hierarchical structure of the documents and incorporates it into an extractive summarization model through a multi-task learning approach. We then present our results on three long summarization datasets, arXiv-Long, PubMed-Long, and Longsumm. Our method outperforms or matches the performance of strong baselines. Furthermore, we perform a comprehensive analysis over the generated results, shedding insights on future research for long-form summary generation task. Our analysis shows that our multi-tasking approach can adjust extraction probability distribution to the favor of summary-worthy sentences across diverse sections. Our datasets, and codes are publicly available at \url{https://github.com/Georgetown-IR-Lab/ExtendedSumm}.

\end{abstract}

\section{Introduction}


In the past few years, there has been a significant progress on both extractive \citep[e.g.,][]{Nallapati2017SummaRuNNerAR, Zhou2018NeuralDS, Liu2019TextSW, Xu2020DiscourseAwareNE, Jia2020NeuralES} and abstractive \citep[e.g.,][]{See2017GetTT, Cohan2018ADA, MacAvaney2019SIG, zhang2019pegasus, Sotudeh2020AttendTM, Dong2020MultiFactCI} approaches for document summarization. These approaches generate a concise summary of a document, capturing its salient content. 
However, for a longer document containing numerous details, it is sometimes helpful to read an extended summary, providing details about its different aspects. Scientific papers are examples of such documents; while their abstracts provide a short summary about their main methods and findings, the abstract does not include details of the methods or experimental conditions. To those who seek more detailed information about a document without having to cover the entire document, an extended or long summary can be desirable~\cite{chandrasekaran-etal-2020-overview-insights, sotudeh-gharebagh-etal-2020-guir, ghosh-roy-etal-2020-summaformers}.
Many long documents, including scientific papers, follow a certain hierarchical structure where content is organized throughout multiple sections and sub-sections. For example, research papers often describe objectives, problem, methodology, experiments, and conclusions \cite{collins-etal-2017-supervised}. Few prior studies have noted the importance of documents' structure in shorter-form summary generation \cite{collins-etal-2017-supervised, Cohan2018ADA}. However, we are not aware of existing summarization methods explicitly approaching modeling the document structure when it comes to generating \emph{extended summaries}.

We approach the problem of generating extended summary by incorporating document's hierarchical structure into the summarization model. Specifically, we hypothesize that integrating the processes of sentence selection and section prediction improves the summarization model's performance over the existing baseline models on extended summarization task. To substantiate our hypothesis, we test our proposed model on three extended summarization datasets, namely, arXiv-Long, PubMed-Long, and Longsumm. We further provide comprehensive analyses over the generated results for two long datasets, demonstrating the qualities of our model over the baseline. Our analysis reveals that the multi-tasking model helps with adjusting sentence extraction probability to the advantage of salient sentences scattered across different sections of the document. Our contributions are threefold:
\begin{enumerate}
    \item A multi-task learning approach for leveraging document structure in generating extended summaries of long documents.
    \item  In-depth and comprehensive analyses over the generated results to explore the qualities of our model in comparison with the baseline model.
    \item  Collecting two large-scale extended summarization datasets with oracle labels for facilitating ongoing research in extended summarization domain.
\end{enumerate}



\section{Related Work}


\subsubsection{Scientific document summarization} Summarizing scientific papers has garnered vast attention from the research community during recent years, although it has been studied for decades. The characteristics of scientific papers, namely the length, writing style, and discourse structure, lead to special model considerations to overcome the summarization task in scientific domain. Researchers have utilized different approaches to address these challenges. 
In earlir work, \citet{Teufel2002SummarizingSA} proposed a Na\"ive bayes classifier to do content selection over the documents' sentences with regard to their rhetorical sentence role.
More recent works have given rise to the importance of discourse structure and its usefulness in summarizing scientific papers. For example, \citet{collins-etal-2017-supervised} used a set of pre-defined section clusters that source sentences are appeared in as a categorical feature to aid the model at identifying summary-worthy sentences. \citet{Cohan2018ADA} introduced large-scale datasets of arXiv and PubMed (collected from public repositories), and used a hierarchical encoder to model the discourse structure of a paper, and then used an attentive decoder to generate the summary. More recently, \citet{Xiao2019ExtractiveSO} proposed a sequence-to-sequence model that incorporates both the global context of the entire document and local context within the specified section. Inspired by the fact that discourse information is important when dealing with long documents \citep{Cohan2018ADA}, we utilize this structure in scientific summarization. Unlike prior works, we integrate sentence selection and sentence section labeling processes through a multi-task learning approach.
In a different line of research, the use of citation context information has been shown to be quite effective at summarizing scientific papers~\cite{AbuJbara2011CoherentCS}. For instance, \citet{Cohan2015ScientificAS,cohan2018scientific} utilized a citation-based approach, denoting how the paper is cited in the reference papers, to form the summary. Here, we do not exploit any citation context information. 



\subsubsection{Extended summarization} While summarization research has been extensively explored in literature, extended summarization has recently gained a huge deal of attention from the research community. Among the first attempts to encourage the ongoing research in this field, \citet{chandrasekaran-etal-2020-overview-insights} set up the Longsumm shared task~\footnote{\url{https://ornlcda.github.io/SDProc/sharedtasks.html}} on producing extended summaries from scientific documents and provided a extended summarization dataset called Longsumm over which participants were urged to generate extended summaries. To tackle this challenge, researchers used different methodologies. For instance, \citet{sotudeh-gharebagh-etal-2020-guir} proposed a multi-tasking approach to jointly learn sentence importance along with its section to be included in the summary. Herein, we aim at validating the multi-tasking model on a variety of extended summarization datasets and provide a comprehensive analysis to guide future research. Moreover, ~\citet{ghosh-roy-etal-2020-summaformers} utilized section-contribution pre-computations (training set) to assign weights via a budget module for generating extended summaries. After specifying the section contribution, an extractive summarizer is executed over each section separately to extract salient sentences. Unlike their work, we unify sentence selection and sentence section prediction tasks to effectively aid the model at identifying summary-worthy sentences scattered around different sections. Furthermore, ~\citet{reddy-etal-2020-iiitbh} proposed a CNN-based classification network for extracting salient sentences. \citet{gidiotis-etal-2020-auth} proposed to use a divide and conquer (DANCER) approach ~\cite{Gidiotis2020ADA} to identify the key sections of the paper to be summarized. The PEGASUS abstractive summarizer~\cite{zhang2019pegasus} then runs over each section separately to produce section summaries, which are finally concatenated to form the extended summary. \citet{Beltagy2020LongformerTL}
 proposed ``Longformer'' that utilizes ``Dilated Sliding Windows'', enabling the model to achieve better long-range coverage on long documents. 
With all being mentioned above, to the best of our knowledge, we are the first to conduct quite a comprehensive analysis over the generated summarization results in the extended summarization domain. 

\section{Dataset}
We use three extended summarization datasets in this research. The first one is Longsumm dataset, which has been provided in the Longsumm 2020 shared task~\cite{chandrasekaran-etal-2020-overview-insights}. To further validate the model, we collect two additional datasets called arXiv-Long and PubMed-Long by filtering the instances of arXiv and PubMed corpora to retain those whose abstract contains at least 350 tokens. Also, to measure how our model works on the mixed varied-length scientific dataset, we exploit the arXiv summarization dataset \cite{Cohan2018ADA}.

\subsubsection{Longsumm} The Longsumm dataset  was provided for the Longsumm challenge ~\cite{chandrasekaran-etal-2020-overview-insights} whose aim was to generate extended summaries for scientific papers. It consists of two types of summaries:
\begin{itemize}
    \item Extractive summaries: these summaries are coming from the TalkSumm dataset \citep{Lev2019TalkSummAD}, containing 1705 extractive summaries of scientific papers according to their video talks in conferences (i.e., ACL, NAACL, etc.). Each summary within this corpus is formed by appending top 30 sentences of the paper.
    \item Abstractive summaries: an add-on dataset containing 531 abstractive summaries from several CS domains such as Machine Learning, NLP, and AI, that are written by NLP and ML researchers on their blogs. The length of summaries in this dataset ranges from 50-1500 words per paper.
\end{itemize}

In our experiments, we use the extractive set along with 50\% of the abstractive set as our training set, containing 1969 papers; and 20\% of it as the validation set. Note that these splits are made for the purpose of our internal experiments as the official test set containing 22 abstractive summaries is blind~\cite{chandrasekaran-etal-2020-overview-insights}.
\begin{figure*}[t]
    \centering
    \includegraphics[scale=0.50]{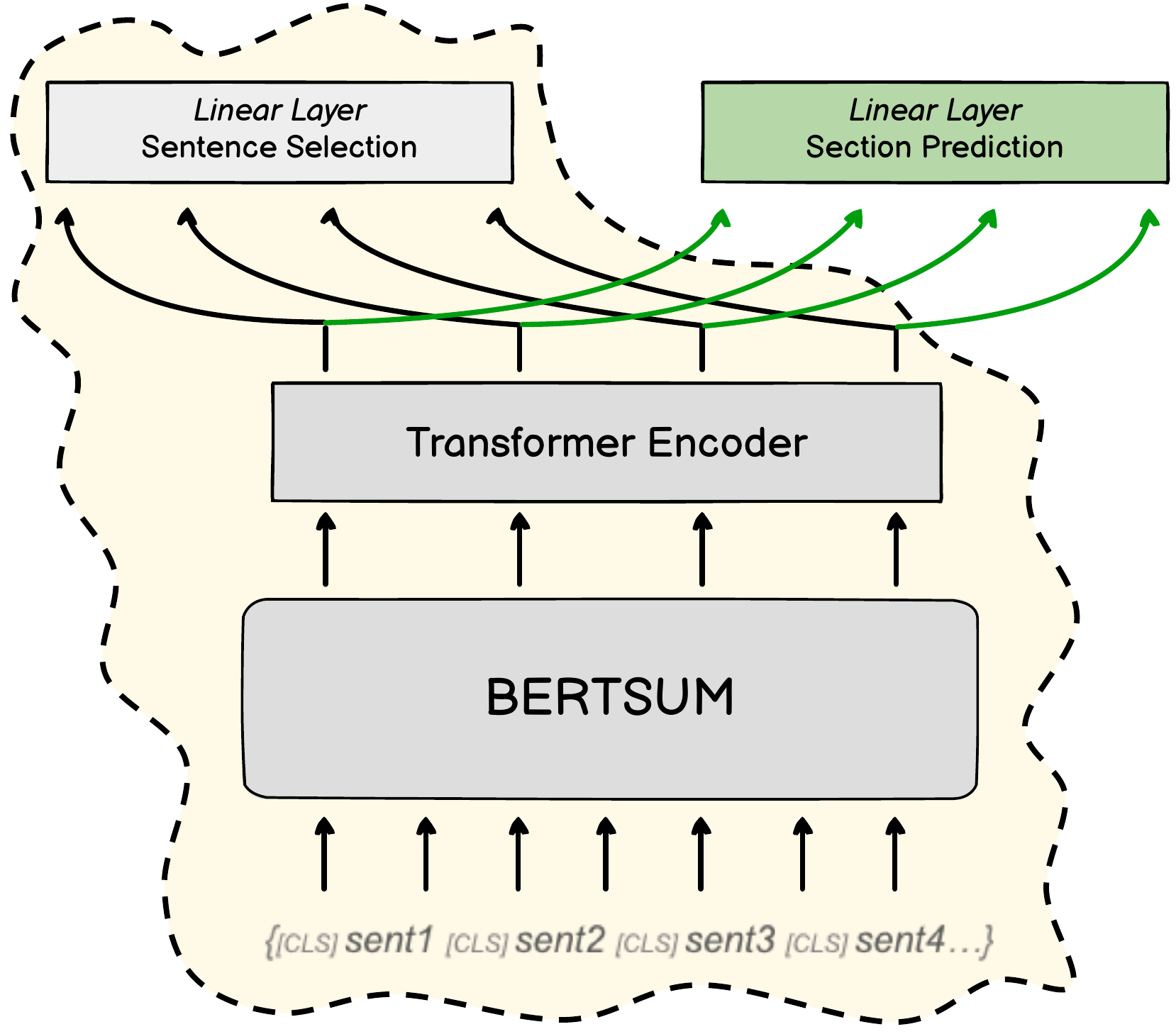}
    \caption{The overview of \textsc{BertSumExtMulti} model. The baseline model (i.e., \textsc{BertSumExt}) is dash-boarded. The extension to the baseline model is addition of Section Prediction linear layer (specified in green box).}
    \label{fig:model}
\end{figure*}

\subsubsection{arXiv-Long \& PubMed-Long.} To further test our methods on additional datasets, we construct two extended summarization datasets for our task. For creating the first dataset, we take arXiv summarization dataset introduced by~\citet{Cohan2018ADA} and filter the instances whose abstract (i.e., ground-truth summary) contains at least 350 tokens. We call this dataset arXiv-Long. We repeat the same process on the PubMed papers obtained from the Open Access FTP service~\footnote{\url{https://www.ncbi.nlm.nih.gov/pmc/tools/ftp}} and call this dataset PubMed-Long. The motivation is that we are interested in validating our model on extended summarization datasets to investigate its effects compared to the existing works, and 350 is the length threshold that we use to characterize papers with ``long'' summaries. The resulting sets contain 11,149 instances for arXiv-Long, and 88,035 instances for PubMed-Long datasets. Note that the abstract of papers are used as ground-truth summaries in these two datasets. The overall statistics of the datasets are shown in Table \ref{tab:dsStat}. We release these datasets to facilitate future research in extended summarization.~\footnote{\url{https://github.com/Georgetown-IR-Lab/ExtendedSumm}} 

\begin{table}[h]

\scalebox{.78}{
\begin{tabular}{lrrr}
\toprule
 \multirow{2}{*}{Datasets} & \multirow{2}{*}{\# docs} & avg. doc. length & avg. summ. length\\
 &  &  (tokens) & (tokens) \\

\midrule\vspace{-1em}
\\
arXiv & 215K & 4938 & 220\\
Longsumm & 2.2K & 5858 & 920\\
arXiv-Long & 11.1K & 9221 & 574 \\
PubMed-Long & 88.0K & 5359 & 403\\

\bottomrule

\end{tabular}
}

\caption{Statistics on arXiv~\cite{Cohan2018ADA},  Longsumm~\cite{chandrasekaran-etal-2020-overview-insights}, and two extended summarization datasets (arXiv-Long, PubMed-Long), collected by this work.}
\label{tab:dsStat}
\end{table}


\vspace{-1em}

\section{Methodology}

In this section, we discuss our proposed method that aims at jointly learning to predict sentence importance and its corresponding section. Before discussing the details of our summarization model, we investigate the preliminary background that provides a fair basis for implementing our method.

\subsection{Background}
\subsubsection{Extractive Summarization}
The extractive summarization system aims at extracting salient sentences to be included in the summary. Formally, let $P$ show a scientific paper containing sentences $[s_1, s_2, s_3, ..., s_m]$, where $m$ is the number of sentences. The extractive summarization is then defined as the task of assigning a binary label ($\hat{y}_i \in \{0,1\}$) to each sentence $s_i$ within the paper, signifying whether the sentence should be included in the summary.

\subsubsection{\textsc{BertSum}: \textsc{Bert} for Summarization}
As our base model we use the \textsc{BertSum} extractive summarization model \cite{Liu2019TextSW}, a \textsc{Bert}-based sentence classification model fine-tuned for summarization. 


After \textsc{BertSum} outputs sentence representations within the input document, several inter-sentence Transformer layers are stacked upon the \textsc{BertSum} to collect document-level features. The final output layer is a linear classifier with Sigmoid activation function to decide whether the sentence should be included or not. The loss function is defined as below:

\begin{equation}
     \mathcal{L}_1 = -\frac{1}{N} \sum_{i=1}^{n} y_i\mbox{log}(\hat{y_i}) + (1-y_i)\mbox{log}(1-\hat{y_i})
 \end{equation}
where $N$ is the output size, $\hat{y_i}$ is the output of the model, and $y_i$ is the corresponding target value. In our experiments, we use this model to extract salient sentences (i.e., those with the positive label) to form the summary. We set this model as the baseline called \textsc{BertSumExt}~\cite{Liu2019TextSW}. 


\subsection{Our model: a section-aware summarizer}

Inspired by few prior works that have studied the effect of document's hierarchical structure in summarization task \cite{Conroy2017SectionMM, Cohan2018ADA}, we define a section prediction task, aiming at predicting the relevant section for each sentence in the document. Specifically, we add an additional linear classification layer on top of \textsc{BertSum} sentence representations to predict the relevant section to each sentence. The loss function for the section prediction network is defined as follows:

\begin{equation}
     \mathcal{L}_2 = -\sum_{i=1}^{S} y_i\mbox{log}(\hat{y_i})
 \end{equation}
where $y_i$ and $\hat{y}_i$ are the ground-truth and the model scores for each section $i$ in $S$.

The entire extractive network is then trained to optimize both tasks (i.e., sentence selection and section prediction) in a multi-task setting:

\begin{equation}
     \mathcal{L}_{\text{Multi}} = \alpha \mathcal{L}_1 + (1-\alpha) \mathcal{L}_2
 \end{equation}
where $\mathcal{L}_1$ is the binary cross-entropy loss from sentence selection task, $\mathcal{L}_2$ is the categorical cross-entropy loss from section prediction network, and $\alpha$ is the weighting parameter that balances the learning procedure between the sentence and section prediction tasks.
\begin{table*}[t]
\centering 
\begin{center}

\begin{tabular*}{\textwidth}{l@{\hspace{3em}}lclllllll}
\toprule

     & & \multicolumn{3}{c}{Validation} &  & & \multicolumn{3}{c}{Test} \\
 \cline{3-5}  \cline{7-9}
 Model                   & Dataset & \small RG-1(\%)  &\small RG-2(\%)  &\small RG-L(\%)  & &\small RG-1(\%)  &\small RG-2(\%)  &\small RG-L(\%) \\

\midrule
 \textsc{BertSumExt}                     &  \multirow{2}{*}{Longsumm}   & \fontsize{11}{60}\selectfont 43.2 &	\fontsize{11}{60}\selectfont 12.4 &	\fontsize{11}{60}\selectfont 16.8  & & \fontsize{11}{60}\selectfont --&	\fontsize{11}{60}\selectfont --&	\fontsize{11}{60}\selectfont -- \\
  \textsc{BertSumExtMulti}              &  & \fontsize{11}{60}\selectfont \textbf{43.3}	& \fontsize{11}{60}\selectfont \textbf{13.0\oneS}&	\fontsize{11}{60}\selectfont \textbf{17.0}  & & \fontsize{11}{60}\selectfont 53.1&	\fontsize{11}{60}\selectfont 16.8&	\fontsize{11}{60}\selectfont 20.3 \\

\midrule
 \textsc{BertSumExt}                     &  \multirow{2}{*}{arXiv-Long}   & \fontsize{11}{60}\selectfont 47.1 &	\fontsize{11}{60}\selectfont 18.2 &	\fontsize{11}{60}\selectfont 20.8  & & \fontsize{11}{60}\selectfont 47.2&	\fontsize{11}{60}\selectfont 18.4&	\fontsize{11}{60}\selectfont 21.1 \\
  \textsc{BertSumExtMulti}              &  & \fontsize{11}{60}\selectfont \textbf{47.8\oneS}	& \fontsize{11}{60}\selectfont \textbf{18.9\oneS}&	\fontsize{11}{60}\selectfont \textbf{21.3\oneS}  & & \fontsize{11}{60}\selectfont \textbf{47.8\oneS}&	\fontsize{11}{60}\selectfont \textbf{19.2\oneS}&	\fontsize{11}{60}\selectfont \textbf{21.5\oneS} \\

\midrule
 \textsc{BertSumExt}                     &  \multirow{2}{*}{PubMed-Long}   & \fontsize{11}{60}\selectfont \textbf{49.1} &	\fontsize{11}{60}\selectfont \textbf{24.3} &	\fontsize{11}{60}\selectfont \textbf{25.7}  & & \fontsize{11}{60}\selectfont \textbf{49.1}&	\fontsize{11}{60}\selectfont \textbf{24.5}&	\fontsize{11}{60}\selectfont \textbf{25.8} \\
  \textsc{BertSumExtMulti}              &  & \fontsize{11}{60}\selectfont 48.9	& \fontsize{11}{60}\selectfont 24.1&	\fontsize{11}{60}\selectfont 25.5  & & \fontsize{11}{60}\ 48.9&	\fontsize{11}{60}\selectfont 24.1&	\fontsize{11}{60}\selectfont 25.5 
  \vspace{0.01em} \\
\bottomrule
\end{tabular*}
\end{center}
\caption{\textsc{Rouge (F1)} results of the baseline (i.e., \textsc{BertSumExt}) and our proposed model (i.e., \textsc{BertSumExtMulti}) on extended summarization datasets. \oneS shows the statistically significant improvement (paired t-test, $p<0.01$). The validation set for Longsumm refers to our internal validation set (20\% of the abstractive set) as there was no official validation set provided for this dataset.}
\label{tab:summ}

\end{table*}

\begin{table*}
\centering 
\begin{center}

\centering 
 \begin{tabular}{ l @{\hspace{5\tabcolsep}} rrrc}
 \toprule
  & \fontsize{10}{60}\selectfont RG-1 & \fontsize{10}{60}\selectfont RG-2  & \fontsize{10}{60}\selectfont RG-L & F-Measure average \\
 \midrule
 \textit{Other systems} \\
\hspace{1em}\fontsize{10.5}{60}\selectfont Summaformers~\cite{ghosh-roy-etal-2020-summaformers} & \fontsize{11}{60}\selectfont 49.38 & \fontsize{11}{60}\selectfont \textbf{16.86} & \fontsize{11}{60}\selectfont \textbf{21.38} & 29.21  \\
\hspace{1em}\fontsize{10.5}{60}\selectfont Wing & \fontsize{11}{60}\selectfont 50.58 & \fontsize{11}{60}\selectfont 16.62 & \fontsize{11}{60}\selectfont 20.50   & 29.23 \\
\hspace{1em}\fontsize{10.5}{60}\selectfont IIITBH-IITP~\cite{reddy-etal-2020-iiitbh} & \fontsize{11}{60}\selectfont 49.03 & \fontsize{11}{60}\selectfont 15.74 & \fontsize{11}{60}\selectfont 20.46 & 28.41 \\
\hspace{1em}\fontsize{10.5}{60}\selectfont Auth-Team~\cite{gidiotis-etal-2020-auth} & \fontsize{11}{60}\selectfont 50.11 & \fontsize{11}{60}\selectfont 15.37 & \fontsize{11}{60}\selectfont 19.59 & 28.36 \\
\hspace{1em}\fontsize{10.5}{60}\selectfont CIST\_BUPT~\cite{li-etal-2020-cist} & \fontsize{11}{60}\selectfont 48.99 & \fontsize{11}{60}\selectfont 15.06 & \fontsize{11}{60}\selectfont 20.13 & \fontsize{11}{60}\selectfont 28.06 \\
 \midrule
 \textit{This work} \\
 \hspace{1em} \fontsize{10.5}{60}\selectfont \textsc{BertSumExtMulti} & \fontsize{11}{60}\selectfont \textbf{53.11} & \fontsize{11}{60}\selectfont 16.77 & \fontsize{11}{60}\selectfont 20.34 & \textbf{30.07} \\

\bottomrule
\end{tabular}

\caption{\textsc{Rouge (F1)} results of our multi-tasking model on the blind test set of Longsumm shared task containing 22 abstractive summaries \cite{chandrasekaran-etal-2020-overview-insights}, along with the performance of other participants' systems. We only show top 5 participants in this table.}
\label{tab:blind}
\end{center}
\end{table*}


\begin{table*}[t]
\centering 
\begin{center}
\begin{tabular*}{\textwidth}{l@{\hspace{4em}}l@{\hspace{4.5em}}rrrrrrrr}
\toprule

     & & \multicolumn{3}{c}{Validation} &  & & \multicolumn{3}{c}{Test} \\
 \cline{3-5}  \cline{7-9}
 Model                   & Dataset & \small RG-1(\%)  &\small RG-2(\%)  &\small RG-L(\%)  & &\small RG-1(\%)  &\small RG-2(\%)  &\small RG-L(\%) \\

 \midrule
 \textsc{BertSumExt}                     &  \multirow{2}{*}{arXiv}   & \fontsize{11.5}{60}\selectfont \textbf{43.6} &	\fontsize{11.5}{60}\selectfont \textbf{16.6} &	\fontsize{11.5}{60}\selectfont \textbf{20.2}  & & \fontsize{11.5}{60}\selectfont \textbf{44.0}&	\fontsize{11.5}{60}\selectfont \textbf{16.8}&	\fontsize{11.5}{60}\selectfont \textbf{20.4} \\
  \textsc{BertSumExtMulti}              &  & \fontsize{11.5}{60}\selectfont 43.4	& \fontsize{11.5}{60}\selectfont 16.5&	\fontsize{11.5}{60}\selectfont 19.8  & & \fontsize{11.5}{60}\selectfont 43.5&	\fontsize{11.5}{60}\selectfont 16.5&	\fontsize{11.5}{60}\selectfont 20.0 \\
\bottomrule
\end{tabular*}
\end{center}
\caption{\textsc{Rouge (F1)} results of the baseline (i.e., \textsc{BertSumExt}) and our proposed model (i.e., \textsc{BertSumExtMulti}) on arXiv summarization dataset.}
\label{tab:arxsum}

\end{table*}


\section{Experimental Setup}
In this section, we give details about the pre-processing steps on the datasets and parameters that we used for the experimented models.

For our baseline, we used the pre-trained \textsc{BertSum} model and implementation provided by the authors \cite{Liu2019TextSW}.\footnote{\url{https://github.com/nlpyang/PreSumm}} The \textsc{BertSumExtMulti} is that of the model used in \cite{sotudeh-gharebagh-etal-2020-guir}, but without post-processing module at inference time, which utilizes trigram-blocking~\cite{Liu2019TextSW} to hinder repetitions in the final summary. We intentionally removed the post-processing part as the model could attain higher scores in the absence of this module throughout our experiments. In order to obtain ground-truth section labels associated with each sentence, we utilized the external sequential-sentence package\footnote{\url{https://github.com/allenai/sequential_sentence_classification}} by \citet{Cohan2019PretrainedLM}. To provide oracle labels for source sentences in our datasets, we use a greedy labelling approach \cite{Liu2019TextSW} with slight modification for labelling up top 30, 15, and 25 sentences for Longsumm, arXiv-Long, and PubMed-Long datasets, respectively, since these numbers of oracle sentences yielded the highest oracle scores.~\footnote{The modification was made to assure that the oracle sentences are sampled from diverse sections.} For the joint model, we tuned $\alpha$ (loss weighting parameter) at 0.5 as it resulted in the highest scores throughout our experiments. In all our experiments, we pick the checkpoint that achieves the best average of \textsc{Rouge-2} and \textsc{Rouge-L} scores on the validation intervals as our best model for inference.

\section{Results}
In this section, we present the performance of the baseline and our model over the validation and test sets of the extended summarization datasets.  We then discuss our proposed model's performance compared to baseline over a mix of varied-length summarization dataset (i.e., arXiv). As the evaluation metrics, we report the summarization systems' performance in terms of \textsc{Rouge-1 (F1)}, \textsc{Rouge-2 (F1)}, and \textsc{Rouge-L (F1)}) metrics.


As we see in Table \ref{tab:summ}, we notice that having section predictor model incorporated into summarization model (i.e., \textsc{BertSumExtMulti} model) performs fairly well compared to the baseline model. This is a particularly important finding since it characterizes the importance of injecting documents' structure when summarizing a scientific paper. While the score gap is relatively higher in arXiv-Long and Longsumm datasets, it is similar in PubMed-Long dataset.

As observed in Table. \ref{tab:blind}, it is noticeable that \textsc{BertSumExtMulti} approach performs top among the state-of-the-art long summarization methods on the blind test set of LongSumm challenge~\cite{chandrasekaran-etal-2020-overview-insights}. While this model improves \textsc{Rouge-1} quite significantly over the other state-of-the-art, it stays competitive on \textsc{Rouge-2} and \textsc{Rouge-L} metrics. In terms of \textsc{Rouge} (F1) F-Measure average, \textsc{BertSumExtMulti} model ranks first by a huge margin compared to the other systems.

To test the model on mixed varied-length summarization datasets, we trained and tested it on arXiv~\cite{Cohan2018ADA} dataset, which contains a mix of varying length abstracts as ground-truth summaries. Table \ref{tab:arxsum} shows that our model can achieve competitive performance on this dataset. While the model does not yield any improvement on arXiv dataset, our hypothesis was to investigate if our model is superior to existing models on longer-form datasets --such as those we have used in this research, which we validated by presenting the evaluation results on long summarization datasets.

\section{Analysis}
In order to gain insights into how our multi-tasking approach works on different long datasets, we perform an extensive analysis in this section to explore the qualities of our multi-tasking system (i.e., \textsc{BertSumExtMulti}) over the baseline (i.e., \textsc{BertSumExt}). Specifically, we perform two types of analyses: 1) quantitative analysis; 2) qualitative analysis. 

For the first part, we choose to use two metrics: \rgdiff{} which denotes the average \textsc{Rouge (F1)} difference (i.e., gap) between the baseline and our model~\footnote{The average is defined on \textsc{Rouge-1 (F1)}, \textsc{Rouge-2 (F1)}, and \textsc{Rouge-L (F1)} scores.}. Positive values indicate the improvement, while negative values denote the decline in scores. Similarly, \fdiff{} is the average difference of F1 score between the baseline and our model. We create three bins sorted by \rgdiff{}: \textsc{Improved} which contains the reports whose average \textsc{Rouge (F1)} score is improved by the multi-tasking model; \textsc{Tied} including those that the multi-tasking model leaves unchanged in terms of modifying average \textsc{Rouge (F1)} score; and \textsc{Declined} containing those whose average \textsc{Rouge (F1)} score has decreased by the joint model.

\begin{figure*}
\centering
\bgroup
\setlength{\tabcolsep}{0pt}
\renewcommand{\arraystretch}{0}
\begin{tabular}{c@{\hspace{5em}}c}

\includegraphics[scale=0.47]{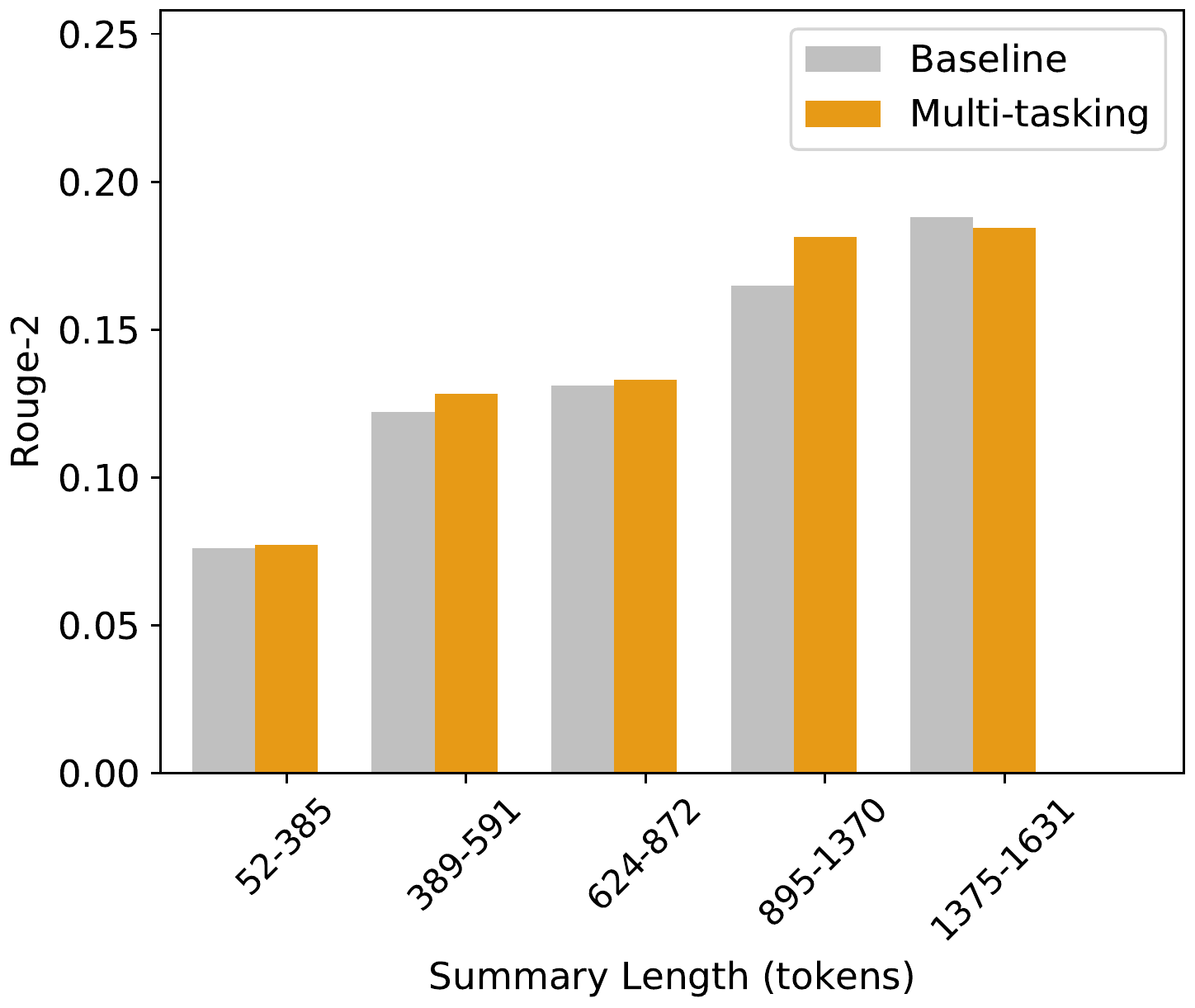} & 
\includegraphics[scale=0.47]{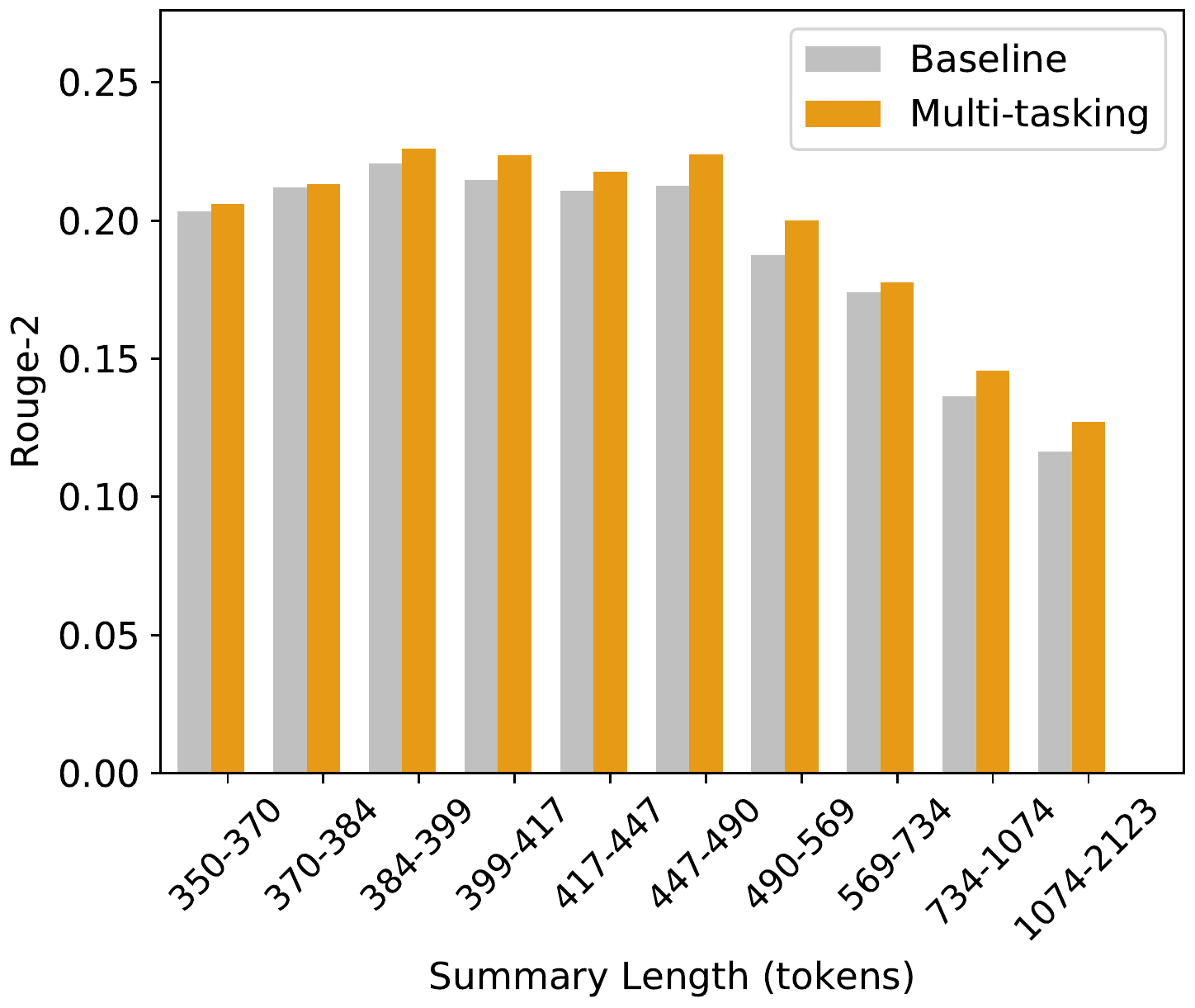}\vspace{0.5em}\\
 \vspace{0.5em}
(a) \textsc{Rouge-2} scores on Longsumm & (b) \textsc{Rouge-2} scores on arXiv-Long \\
\end{tabular}
\caption{Bar charts exhibiting the correlation of ground-truth summary length (in tokens) with the baseline (i.e., \textsc{BertSumExt}) and our multi-tasking model's (i.e., \textsc{BertSumExtMulti}) performance. The diagrams are shown for Longsumm and arXiv-Long  datasets' test set. Each bin contains 31 summaries for Longsumm, and 196 summaries for arXiv-Long. As denoted, the multi-tasking model generally outperforms the baseline on later bins which include longer-form summaries.}
\label{fig:longsumm}
\egroup
\end{figure*}

\begin{figure*}[t]
\centering
\bgroup
\setlength{\tabcolsep}{0pt}
\renewcommand{\arraystretch}{0}
\begin{tabular}{c}
\includegraphics[scale=0.053]{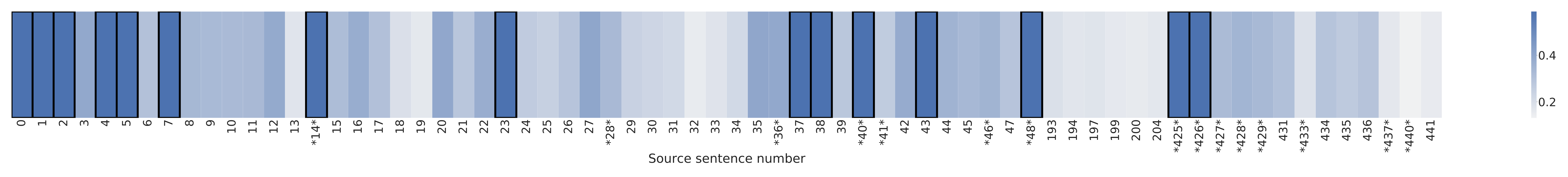} \\
\multicolumn{1}{l}{\hspace{0.6cm}(a) Extraction probability distribution of the baseline model (i.e., \textsc{BertSumExt}) over the source sentences.} \\
\includegraphics[scale=0.053]{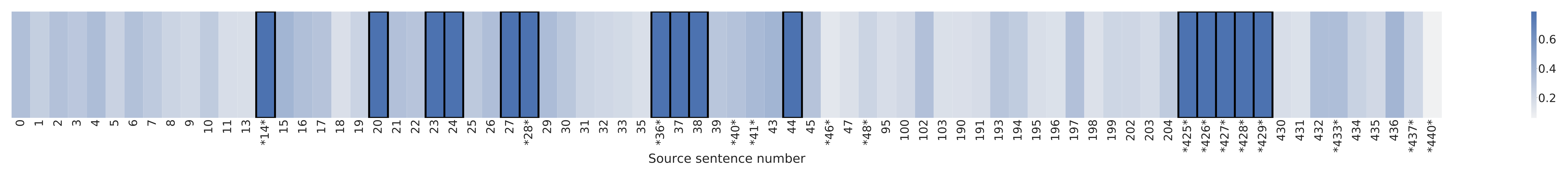} \\
\multicolumn{1}{l}{\hspace{-0.2cm}(b) Extraction probability distribution of the multi-tasking model (i.e., \textsc{BertSumExtMulti}) over the source sentences.} \\
\end{tabular}
\vspace{-0.5em}
\caption{Heat-maps showing the extraction probabilities over the source sentences (Paper ID: \texttt{astro-ph9807040} sampled from arXiv-Long dataset). For simplicity, we have only shown the sentences that gain over 15\% extraction probability by the models. The cells bordered in black show the models' final selection, and oracle sentences are indicated with *.}
\label{fig:heatmap}
\egroup
\end{figure*}


For the qualitative analysis section, we specifically aim at comparing the methods in terms of section distribution since that is where our method's improvements are expected to come from. Furthermore, we conduct an additional length analysis over the results generated by the baseline versus our model.

\begin{table}[h]
    \centering
    
    \scalebox{0.9}
    {
    \begin{tabular}{lcrrr}
    \toprule
        Bin & Dataset & Count & \rgdiff{} & \fdiff{}  \\
         
        \midrule
                        \vspace{-0.4em} \\

       \textsc{Improved} & \multirow{3}{*}{Longsumm} & 76 & 2.05 & 6.16 \\
       \textsc{Tied} & & 4 & 0 & 0 \\
       \textsc{Declined} & & 74 & \textminus1.47 & 1.95\vspace{0.5em}\\
        \midrule
        Total & & 154 & 0.31 & 4.11 \\
        \bottomrule
                \vspace{-0.4em} \\

        \textsc{Improved} & \multirow{3}{*}{arXiv-Long} & 1,084 & 2.40 & 4.47 \\
       \textsc{Tied} & & 67 & 0 & 0.32 \\
       \textsc{Declined} &  & 801 & \textminus1.82 & \textminus1.34\vspace{0.5em}  \\
        \midrule
        Total & & 1952 & 0.59 & 1.94 \\
        \bottomrule
    \end{tabular} 

    }
    \vspace{0.1cm}
    \caption{\textsc{Improved}, \textsc{Tied}, and \textsc{Declined} bins on the test set of Longsumm and arXiv-Long datasets. The numbers show the improvements (positive) and drops (negative) compared to the baseline model (i.e., \textsc{BertSumExt}).}
    \label{tab:lsum-ann}
\end{table}

\subsection{Quantitative Analysis}
We first perform the quantitative analysis over the long summarization datasets' test sets in two parts including 1) \textit{Metric analysis} which aims at comparing different bins based on the average \textsc{Rouge} score difference of the baseline and our model; 2) \textit{Length analysis} that targets at finding the correlation between the summary length on different bins and models' performance.

\subsubsection{Metric analysis}
Table \ref{tab:lsum-ann} shows the overall quantities of Longsumm and arXiv-Long datasets in terms of average difference of \textsc{Rouge} and F1 scores. As shown, the multi-tasking approach is able to improve 76 summaries with an average \textsc{Rouge (F1)} improvement of 2.05\%. This is even more when it comes to evaluating the model on arXiv-Long dataset with average \textsc{Rouge} improvement of 2.40\%. 

Interestingly, our method can consistently improve F1 measure in general (See total F1 scores in Table. \ref{tab:lsum-ann}). Seemingly, F1 metric directly correlates with \textsc{Rouge (F1)} metric on arXiv-Long dataset, whereas this is not the case on \textsc{Declined} bin of the Longsumm dataset. This might be due to the relatively small test set size of Longsumm dataset. It has to be mentioned that \textsc{Improved} bin holds relatively higher counts and improved metrics than that of \textsc{Declined} bin across both datasets in our evaluation.

        

\subsubsection{Length analysis} We analyze the generated results by both models to see if the summary length affects the models' performance using bar charts in Figure \ref{fig:longsumm}. The bar charts are intended to provide the basis for comparing both models on different length bins (x-axis), which are evenly-spaced (i.e., having the same number of papers). It has to be mentioned that we used five bins (each bin with 31 summaries) and ten bins (each bin with 196 summaries) for Longsumm and arXiv-Long datasets, respectively.

As shown in Figure \ref{fig:longsumm} (a), for Longsumm dataset, as the length of the ground-truth summary increases, the multi-tasking model generally improves over the baseline consistently on both datasets, except for the last bin on Longsumm dataset where it achieves comparable performance. This behaviour is also observed on \textsc{Rouge-1} and \textsc{Rouge-L} for Longsumm dataset. The \textsc{Rouge} improvement is even more noticeable when it comes to analysis over arXiv-Long dataset (See Figure \ref{fig:longsumm} (b)). Thus, the length analysis supports our hypothesis that the multi-tasking model outperforms the baseline more significantly when the summary is of longer-form.

\subsection{Qualitative analysis}
As the results of the qualitative analysis on the \textsc{Improved} bin is observed, we found out that the multi-tasking model can effectively sample sentences from diverse sections when the ground-truth summary is also sampled from diverse sections. It improves significantly over the baseline when the extractive model can detect salient sentences from important sections. 

By investigating the summaries from\textsc{ Declined} bin, we noticed that in declined summaries, while our multi-tasking approach can adjust extraction probability distribution to diverse sections, it has difficulty picking up salient sentences (i.e., positive sentences) from the corresponding section; thus, it leads to relatively lower \textsc{Rouge} score. This might be improved if two networks (i.e., sentence selection and section prediction) are optimized in a more elegant way such that the extractive summarizer can further select salient sentences from the specified sections when they could be identified. For example, the improved multi-tasking methods can involve task prioritization \cite{Guo2018DynamicTP} to dynamically balance the learning process between two tasks during training, rather than using a fixed $\alpha$ parameter.

In the cases where the F1 score and \textsc{Rouge (F1)} were not consistent with each other, we observed that adding non-salience sentences to the final summary hurts the final \textsc{Rouge (F1)} scores. In other words, while the multi-tasking approach can achieve a higher F1 score compared to the baseline since it chooses different non-salient (i.e., negative) sentences than baseline, the overall \textsc{Rouge (F1)} scores drop slightly. Having conditional decoding length (i.e., sentences) might help with this as done in \cite{Mao2020MultidocumentSW}.

Fig. \ref{fig:heatmap} shows the extraction probabilities that each model outputs on the source sentences. It is observable that the baseline model picks most of the sentences (47\%) from the beginning of the paper, while the multi-tasking approach (b) can effectively distract probability distribution to summary-worthy sentences that are all around different sections of the paper, and pick those with higher confidence. Our model achieves the overall F1 score of 53.33\% on this sample paper, while the baseline's F1 score is 33.33\%.
\section{Conclusion \& Future Work}
In this paper, we approach the problem of generating extended summaries, given a long document. Our proposed model is a multi-task learning approach that unifies sentence selection and section prediction processes, extracting summary-worthy sentences. We further collect two large-scale extended summary datasets (arXiv-Long and PubMed-Long) from scientific papers. Our results on three datasets show the efficacy of the joint multi-task model in the extended summarization task. While it achieves fairly competitive performance with the baseline on one of three datasets, it consistently improves over the baseline in the other two evaluation datasets. We further performed extensive quantitative and qualitative analyses over the generated results by both models. These evaluations revealed our model's qualities compared to the baseline. Based on the error analysis, it could be noticed that the performance of this model highly depends on the multi-tasking objectives. Future studies could fruitfully explore this issue further by optimizing the multi-task objectives in a way that both sentence selection and section prediction tasks can benefit.

\bibliography{emnlp2020}


\end{document}